\documentclass{article}
\usepackage{ijcai19}

\usepackage{times}
\usepackage{soul}
\usepackage{url}
\usepackage[hidelinks]{hyperref}
\usepackage[utf8]{inputenc}
\usepackage[small]{caption}
\usepackage{graphicx}
\usepackage{amsmath}
\usepackage{booktabs}
\usepackage{algorithm}
\usepackage{algorithmic}
\title{3D Graph Embedding Learning with a Structure-aware Loss Function \\
	for Point Cloud Semantic Instance Segmentation}

\author{
Zhidong Liang$^1$
\and
Ming Yang$^2$\And
Chunxiang Wang$^2$
\affiliations
$^1$Research Institute of Robotics, Shanghai Jiao Tong University, China\\
$^2$Department of Automation, Shanghai Jiao Tong University, China
\emails
\{lzd950512\}@sjtu.edu.cn,
}
\begin{document}

\maketitle

\begin{abstract}
  This paper introduces a novel approach for 3D semantic instance segmentation on point clouds. A 3D convolutional neural network called submanifold sparse convolutional network is used to generate semantic predictions and instance embeddings simultaneously. To obtain discriminative embeddings for each 3D instance, a structure-aware loss function is proposed which considers both the structure information and the embedding information. To get more consistent embeddings for each 3D instance, attention-based k nearest neighbour (KNN) is proposed to assign different weights for different neighbours. Based on the attention-based KNN, we add a graph convolutional network after the sparse convolutional network to get refined embeddings. Our network can be trained end-to-end. A simple mean-shift algorithm is utilized to cluster refined embeddings to get final instance predictions. As a result, our framework can output both the semantic prediction and the instance prediction. Experiments show that our approach outperforms all state-of-art methods on ScanNet benchmark and NYUv2 dataset.
\end{abstract}

\section{Introduction}

Recently, semantic instance segmentation is a popular topic in computer vision. As the development of 3D sensors such as RGBD cameras or LIDAR, 3D scene understanding becomes more and more important in augmented reality and autonomous driving. Compared to 2D scene understanding, 3D understanding is more challenging for the data sparsity and the expensive computation cost. However, 3D data contains geometric information which is useful for semantic understanding while 2D images do not. In this paper, we focus on 3D semantic instance segmentation.

Instance segmentation in 2D images achieves a great performance. Most approaches to 2D instance segmentation are proposal-based which first apply a object detector~\cite{girshick2015fast,ren2015faster,redmon2016you,liu2016ssd,lin2017feature,lin2018focal} to get initial bounding boxes and then segment each bounding box binarily to get the instance mask. Such idea~\cite{he2017mask} achieves excellent results which benefits from accurate object detection. However, these methods have some drawbacks. First, they are the combination of object detection and semantic segmentation so the training process is complex. Second, one pixel may have more than one instance labels as they may be in two overlapped bounding boxes simultaneously. The second problem can be more serious when it is the multi-class instance segmentation of clutter scenes.

\begin{figure}[t]
	\centering
	\includegraphics[width=0.95\linewidth]{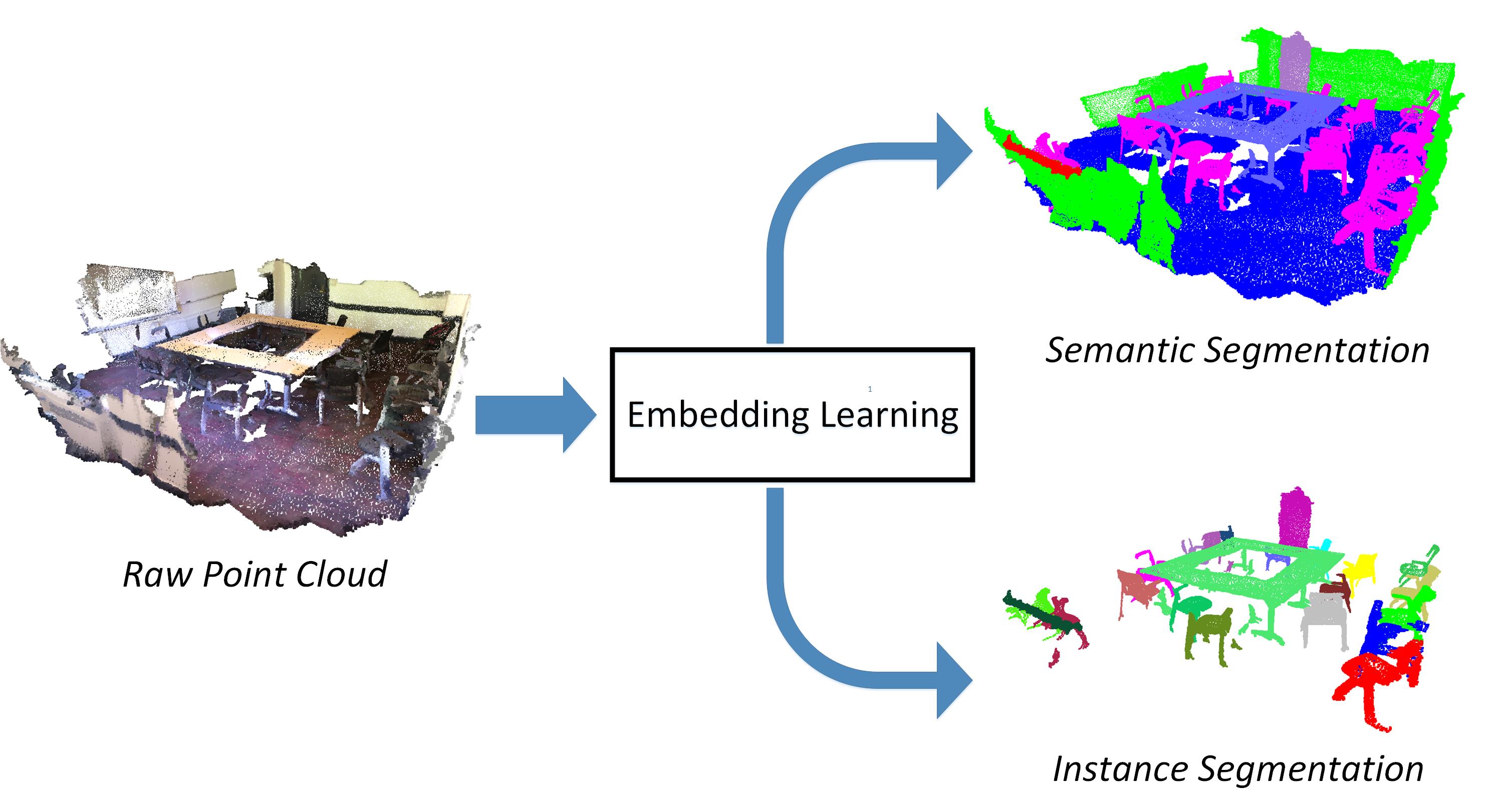}
	\caption{The input of our network is point clouds with xyz coordinates and RGB attributes. The output includes two parts: semantic preditions and instance embeddings. Mean-shift algorithm is used to cluster embeddings to get the final instance predictions.}
	\label{intro}
\end{figure}

An alternative idea is to generate embeddings~\cite{fathi2017semantic,kong2018recurrent,de2017semantic} for each pixel and then apply a cluster algorithm to get the final instance result. This idea can utilize semantic segmentation networks to generate discriminative embeddings. Although such proposal-free methods cannot get as high performance as proposal-based methods in 2D images, they are simpler in the implementation and can avoid the drawbacks of proposal-based methods. Additionally, such framewrok can simultaneously segment images in the semantic level and instance level while the proposal-based method can only get the instance result. It means that objects without the instance label such as the sky or the road may be dropped by the proposal-based method. In this paper, we propose a proposal-free framework for 3D semantic instance segmentation. Figure~\ref{intro} shows the input and the output of our method.

Our backbone network can be arbitary 3D neural network. In this paper, we choose the submanifold sparse convolutional neural network~\cite{graham20183d} to get semantic labels and generate inital embeddings for points. Within an object instance, embeddings of points in the center of the object are more likely to be similar while embeddings near the edge are more likely to be different. To get more consistent embeddings for the same instance, we considering geometric information for 3D instances and propose a structure-aware loss function for 3D instance segmentation. 

In 2D images, adjacent pixels may be far away from each other. Compared to the 2D situation, adjacent points in the 3D space are more likely to be in the same instance. So k nearest neighbour (KNN) algorithm can be used to pass and aggregate information. Message passed from neighbour points cannot only enforce the consistency of embeddings but also eliminate the quantitative error caused by 3D voxel. However, a point and its neighbours are also likely to be in the different instances if the point is near the edge. If so, wrong information will be passed to the point. Considering this problem, we propose an attention-based graph convolutional neural network which automaticly aggregates useful information from neighbours.

The main contributions of this paper are as follows:

\begin{itemize}
	
	\item We propose a structure-aware loss function for 3D instance segmentation which considers both the geometric information and the embedding information for each 3D instance.
	
	\item We propose an attention-based graph convolutional neural network which can automaticly choose and aggregate information from neighbours.
	
	\item We propose a novel method for 3D semantic instance segmentation. Experiments show that our proposed method outperforms all state-of-the-art methods on ScanNet benchmark~\cite{dai2017scannet} and NYUv2 dataset~\cite{silberman2012indoor}.

\end{itemize}

\section{RELATED WORK}

\subsection{Instance Segmentation}
\textbf{Instance Segmentation in 2D Images. }CNN based methods have achieved excellent results on object detection~\cite{girshick2015fast,ren2015faster,redmon2016you,liu2016ssd,lin2017feature,lin2018focal} in 2D images. As a combination of object detection and semantic segmentation, instance segmentation~\cite{he2017mask,fathi2017semantic,kong2018recurrent,de2017semantic} becomes a hot topic in research because it can provide richer semantic information. Inspired by object detection, many approaches to instance segmentation segment the bounding box to get the instance mask. Another idea is to generate an embedding for each pixel and cluster according to the similarity between pixels. Such "proposal-free" methods can avoid some limitations of proposal-based methods. \cite{fathi2017semantic} uses Euclidean distance with a sigmoid function to measure the similarity of each pair of embedding vectors. \cite{kong2018recurrent} utilizes the cosine similarity which is invariant to the scale of the embedding vector. \cite{de2017semantic} proposes a discriminative loss function to pull pixels belonging to the same instance closer in the embedding space.

\textbf{Instance Segmentation in Point Clouds. }Recently, several researchers have tried instance segmentation on point clouds. \cite{wang2018sgpn} is a pioneer in 3D instance segmentation. It generates an embedding for each point and proposes a double-hinge loss to supervise the embedding learning. \cite{yi2018gspn} generates a proposal for each object by reconstructing the shape and then combine PointNet++ to get the final instance segmentation result. \cite{hou20183d} proposes a detection-based method to get the instance prediction which also fuses multi-modal inputs.

\subsection{Deep Learning on Point Clouds}
Deep learning on point clouds develops fast in the recent years. Voxel-based method~\cite{tchapmi2017segcloud} is a natural generalization of 2D convolution. However, the performance of voxel-based method is limited by the resolution of voxels. \cite{riegler2017octnet,wang2017cnn,graham20183d} exploit the sparsity property of 3D data and enable 3D CNN to achieve a higher resolution and efficiency. Additionally, such sparse convolutional operations can be easily combined with many great network frameworks of 2D images. PointNet~\cite{qi2017pointnet} provides a brand new direction for 3D deep learning. It directly process raw point clouds without quantitive errors. 

Recently, graph CNN~\cite{bruna2013spectral,kipf2016semi,simonovsky2017dynamic,wang2018dynamic} is popular in research as it can process unregular data. It can be viewed as the generalization of conventional convolution operation in the non-Euclidean space. Graph convolution has the strong ability to pass messages between neighbours. Point cloud is a kind of graphs which can utilize graph convolution to extract local information using graph convolution. K nearest neighbour algorithm is widely used to search neighbours for point clouds. However, it treats each neighbours equivalently and may bring some improper information.

\section{METHOD DESCRIPTION}
In this section, we first describe the whole network architecture~(Section~\ref{Network_section}). Then we introduce our proposed structure-aware loss function~(Section~\ref{Structure_section}) for supervising the learning of instance embeddings and the attention-based KNN~(Section~\ref{Attention_section}, Section~\ref{Graph_section}) for message aggregation. 
%Finally, the graph convolution based on the attention KNN is used to refine the embedding. We use the submanifold sparse convolutional network as our backbone network. We recommend \cite{graham20183d} for more details on this backbone network. The whole framework of our model is illustrated in Figure \ref{framework}. The input is the raw point cloud with rgb attributes. After the backbone network, we can get initial embeddings for points.

\begin{figure*}[!t]
	\centering
	\includegraphics[width=1.0\linewidth]{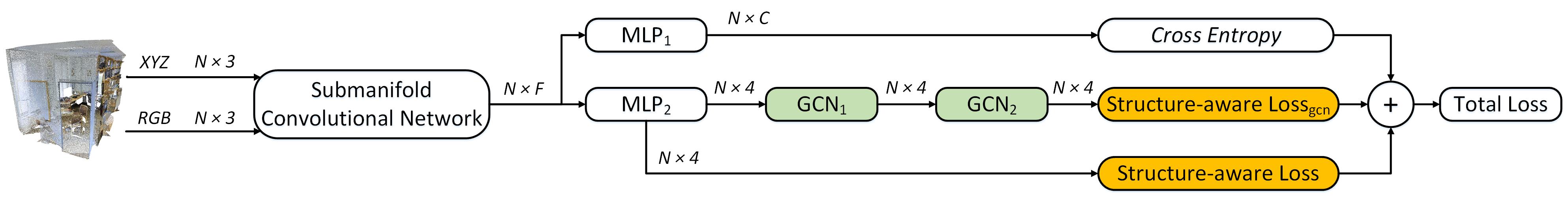}
	\caption{Illustration of the whole network architecture. The input is the original point coordinate with RGB attributes. The output of the submanifold network is the initial embedding for each point. Then two MLPs are followed to generate the semantic prediction and the instance embedding respectively. The number of the semantic class is C and the dimension of the instance embedding is 4 in our paper. Two GCNs are used to refine the instance embedding. The embedding generated by $MLP_{2}$ and the refined embedding optimized by GCNs are both used to calculate the structure-aware loss respectively. The final training process is a multi-task learning. }
	\label{framework}
\end{figure*}

\subsection{Network Architecture} \label{Network_section}
The whole network (illstruted in Figure~\ref{framework}) consists three main components including the submanifold convolutional network, the structure-aware loss function and the graph convolutional network. The architecture of the submanifold convolutional network is borrowed from~\cite{graham20183d}. We recommend \cite{graham20183d} for more details on the network. The ouput of the submanifold convolutional network is the input to two different MLP networks. The first MLP outputs the semantic predictions for the cross entropy loss function. The second MLP outputs the instance embeddings for the structure-aware loss function. To refine the instance embeddings, the output of the second MLP is also inputted to a series of GCNs~(Section~\ref{Graph_section}). Finally, three loss items are added as the total loss.

\subsection{Structure-aware Loss Function} \label{Structure_section}
After generating initial embeddings for all points, we hope that points within the same instance have similar embeddings while points from different instances are apart in the embedding space. This is a classical problem in the metric learning. However, for the 3D instance segmentation task, points withinc each instance do not only have embedding features but also have geometric relations in the 3D space. This is different from the past researches in the metric learning. We combine such structure information with embedding features to make the final results more discriminative.

First, we need to define a metric for measuring the similarity between embeddings. Euclidean distance and cosine distance are commonly used. Cosine distance is scale-invariant to the length of the embedding vector which is an advantage compared to Euclidean distance. However, if using cosine distance, all embeddings need to be lied on a hyper-sphere. One difficulty for instance segmentation is that the number of instance is not uncertain. If the number is too large, the embedding of different instances may be not discriminative enough as they are limited in a hyper-sphere. One solution is to set the dimension of the embedding very high which may cause the learning process and the post-process more difficult. Considering the above reason, we choose the Euclidean distance to measure the similarity for its simplicity.

%SGPN uses a similarity matrix to measure the similarity between each pair of embeddings. The memory allocation of such operation grows quadratically as point number increases. The center loss

We want to minimize the distance between embeddings within the same instance. A mean embedding can be used to describe the overall feature of a instance. For the $i_{th}$ instance, the loss function is formalized as follows:
\begin{equation}
Loss_{i}^{intra}=\sum_{j=1}^{N_{i}} \frac{1}{1+\exp(-p_{i,j})} [s_{i,j}-\alpha]_{+}^{2} \label{intra}
\end{equation}
where $\alpha$ is a threshold for penalizing large embedding distance. $N_{i}$ is the point number of the $i_{th}$ instance. $p_{i,j}$ measures the spatial distance between the $j_{th}$ point and the geometric center $\mu_{p,i}$ of the $i_{th}$ instance and $s_{i,j}$ measures the embedding distance between the $j_{th}$ point and the mean embedding $\mu_{s,i}$:
\begin{equation}
p_{i,j} = \Vert p_{j}-\mu_{p,i} \Vert, \mu_{p,i}=\frac{1}{N_{i}} \sum_{j=1}^{N_{i}} p_{i,j}  \label{spatial}
\end{equation}
\begin{equation}
s_{i,j} = \Vert s_{j}-\mu_{s,i} \Vert, \mu_{s,i}=\frac{1}{N_{i}} \sum_{j=1}^{N_{i}} s_{i,j}  \label{embedding}
\end{equation}
where $p_{i,j}$ and $s_{i,j}$ are the coordinate and the embedding of the $j_{th}$ point within the $i_{th}$ instance.

On the other hand, to make points from different instances discriminative, mean embeddings between different instances should be far away from each other. This idea is commonly used in the previous research in the metric learning:
\begin{equation}
Loss_{ij}^{inter}= [\beta - \Vert \mu_{s,i}-\mu_{s,j} \Vert]_{+}^{2}   \label{inter}
\end{equation}
where $\beta$ is a threshold for the distance between mean embeddings. It means that the loss function just penalizes small distance. If the distance is larger than the threshold, it will not contribute to the loss value as the embeddings are apart enough in the embedding space.

Our final loss function is composed of the above items:
\begin{equation}
Loss=  \frac{1}{M}\sum_{i=1}^{M}Loss_{i}^{intra} + \frac{1}{M(M-1)}\sum_{i=1}^{M}\sum_{j=1,j\not=i}^{M} Loss_{ij}^{inter} \label{strucLoss}
\end{equation}
where $M$ is the total number of the instance in the scene.

\begin{figure*}[!t]
	\centering
	\includegraphics[width=1.0\linewidth]{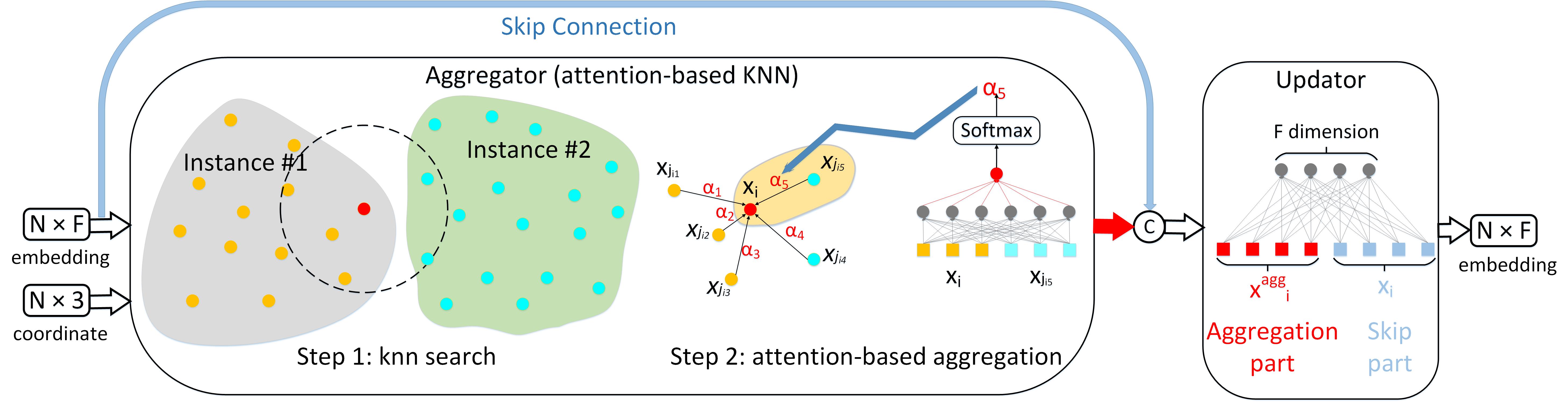}
	\caption{Illustration of the graph convolutional neural network using attention-based KNN. The aggregator is our proposed attention-based KNN~(Section~\ref{Attention_section}). In step one, for each input point, k nearest neighbours are searched according to the spatial coordinate. In step two, different weights are assigned to different neighbours. The output of the aggregator is the weighted average of the embeddings of k neighbours. The skip connection is used to concatenate the ouput of the aggregator and the input embedding. Finally, a fully connected layer is used to get the refined embedding.}
	\label{attention}
\end{figure*}

\subsection{Attention-based K Nearest Neighbour} \label{Attention_section}
The goal of our network is to generate similar embeddings within the same instance and discriminative embeddings between different instances. To achieve this goal, k nearest neighbour (KNN) algorithm can be utilized to enhance the local consistency of embeddings. A point can aggregate information from surrounding points. However, KNN may bring some wrong information which is harmful for embeddings. For example, a point near the edge of a instance may aggregate information from another instances. So we propose an attention-based KNN for embedding aggregation.

The input embeddings of point clouds are denoted by $X=\left\{x_{1},...,x_{n}\right\}$ $\subseteq$ $R^{F}$. $\left\{x_{j_{i_{1}}},...,x_{j_{i_{k}}}\right\}$ are the k nearest neighbours of $x_{i}$ according to their spatial positions. We argue that our KNN use the spatial distance of points instead of the embedding distance as the metric. The standard KNN aggregation process can be formalized as follows:
\begin{equation}
x^{aggregate}_{i}=\frac{1}{k}\sum_{m=1}^kx_{j_{i_{m}}} \label{knn_aggregation}
\end{equation}

To achieve the goal of automatic embedding selection, we utilize the attention mechanism. The operation can be formalized as follows:
\begin{equation}
x^{aggregate}_{i}=\sum_{m=1}^k \alpha_{m}x_{j_{i_{m}}} \label{knn_attention_aggregation}
\end{equation}
where $\alpha_{m}$ is the attention weight for each neighbour. It is related to the embedding of the neighbour and the corresponding center point. It can be calculated as follows:
\begin{equation}
%p_{m}=\left[\begin{array}{c|c} x_{i} & x_{j_{i_{m}}}\end{array}\right] \cdot w \label{attention_weight}
p_{m}=f(x_{i}, x_{j_{i_{m}}}) \label{attention_weight}
\end{equation}
where $f: R^{2 \times F} \mapsto R^{1}$ is a trainable MLP. $\alpha_{m}$ is the normalization of $p_{m}$ using the softmax function:
\begin{equation}
\alpha_{m}=softmax(p_{m})=\frac{\exp(p_{m})}{\sum_{m=1}^k\exp(p_{m})} \label{softmax_weight}
\end{equation}

Compared to the standard KNN aggregation, attention-based KNN can assign different weights for different neighbours. The aggregator in Figure~\ref{attention} illustrates the process of the attention-based KNN. In theory, the network can learn better aggregation strategy than simple average aggregation. Experiments also show its effectiveness.

\subsection{Graph Convolutional Neural Network using Attention-based KNN} \label{Graph_section}
Normally, graph convolutional neural network consists two parts: the aggregator and the updator. The aggregator is to gather information from neighbours and the updator is to update the aggregated information by mapping embeddings into a new feature space. Here, we use our proposed attention-based KNN as the aggregator and use a simple fully connected layer without bias as the updator (illustrated in Figure~\ref{attention}). The operation is formalized as follows:
\begin{equation}
x^{update}_{i}=[x_{i}, x^{aggregate}_{i}] W \label{gcn}
\end{equation}
where $W \subseteq R^{2F \times F}$ is a trainable parameter of the updator.

\textbf{Analysis. }Our proposed graph convolutional network does not need to calculate the laplacian matrix and the eigendecomposition which need huge computation cost. This is very important for graph CNN to apply to the point cloud data. 
%As the number of different point clouds is fixed , the dimension of the laplacian matrix is varying. As a result, eigendecomposition-based methods dose not work here. Even point clouds are downsampled to a fixed number, eigendecomposition-based methods still needs huge computation cost which makes it not practical.
Actually, our method can be viewed as one kind of spatial graph convolutional networks which do not need to compute the eigenvalue of the graph. Also, we use the KNN (complexity $\mathcal{O}(n \times k)$) to decribe the relation instead of using the laplacian matrix form (complexity $\mathcal{O}(n^{2})$). However, the two forms are equivalent essentially. If the laplacian matrix is sparse, some sparse tricks can be used to decrease the computation cost.

The main spotlight of our graph convolutional network is that it uses the attention-based KNN as the aggregator. This is a natural and meaningful operation for point clouds. It allows the network to learn different importances for different neighbours.

\section{EXPERIMENTS}
\textbf{Datasets. }
We evaluate our model in two datasets providing 3D instance segmentation labels:
\begin{itemize}
	\item ScanNet~\cite{dai2017scannet}: This datasets contains 1613 3D indoor scans. We follow the official split of 1201 training samples, 300 validation samples and 100 testing samples (without ground truth). The dataset provides a benchmark for several tasks including 3D instance segmentation. It provides images from different views but we only use the point cloud data in our method.
	
	\item NYUv2~\cite{silberman2012indoor}: This dataset contains 1449 single RGBD images. We follow the same preprocessing method as~\cite{wang2018sgpn} and~\cite{yi2018gspn} to get the 3D annotation of point clouds. We follow the standard split of 795 trainging samples and 654 testing samples. 
\end{itemize}

\begin{table*}[t]
	
	\caption{Results on the test set of ScanNet (v2) 3D instance segmentation benchmark. $AP_{0.5}$ is reported in the table.}
	\label{ScanNet_Result}
	
	\begin{center}
		\resizebox{\textwidth}{!}{
			\begin{tabular}{l|cc|c|cccccccccccccccccc}
				\toprule
				Method & image & point cloud & mean & cabinet & bed & chair & sofa & table & door & window & bookshelf & picture & counter & desk & curtain & fridge & shower & toilet & sink & bathtub & other\\
				\midrule
				Mask R-CNN~\cite{he2017mask} & yes & no & 5.8 & 5.3 & 0.2 & 0.2 & 10.7 & 2.0 & 4.5 & 0.6 & 0.0 & \textbf{23.8} & 0.2 & 0.0 & 2.1 & 6.5 & 0.0 & 2.0 & 1.4 & 33.3 & 2.4 \\
				SGPN~\cite{wang2018sgpn} & no & yes & 14.3 & 6.5 & 39.0 & 27.5 & 35.1 & 16.8 & 8.7 & 13.8 & 16.9 & 1.4 & 2.9 & 0.0 & 6.9 & 2.7 & 0.0 & 43.8 & 11.2 & 20.8 & 4.3\\
				MTML & unknown & unknown & 21.1 & 2.7 & 61.4 & 39.0 & 50.0 & 10.5 & 10.0 & 0.3 & \textbf{33.7} & 0.0 & 0.0 & 0.1 & 11.8 & 16.7 & 14.3 & 57.0 & 4.6 & 66.7 & 2.8\\
				3D-BEVIS    & unknown & unknown & 24.8 & 3.5 & 56.6 & 39.4 & 60.4 & 18.1 & 9.9 & 17.1 & 7.6 & 2.5 & 2.7 & 9.8 & 3.5 & 9.8 & 37.5 & 85.4 & 12.6 & 66.7 & 3.0\\
				R-PointNet~\cite{yi2018gspn}  & no & yes & 30.6 & \textbf{34.8} & 40.5 & \textbf{58.9} & 39.6 & 27.5 & 28.3 & 24.5 & 31.1 & 2.8 & 5.4 & 12.6 & 6.8 & 21.9 & 21.4 & 82.1 & 33.1 & 50.0 & 29.0\\
				3D-SIS~\cite{hou20183d}   & yes & yes   & 38.2 & 24.5 & 43.2 & 57.7 & \textbf{69.9} & 27.1 & 32.0 & 23.5 & 24.5 & 7.5 & 1.3 & 3.3 & 26.3 & \textbf{42.2} & \textbf{85.7} & 88.3 & 11.7 & \textbf{100.0} & 24.0 \\
				UNet-backbone(ours)    & no & yes & 31.9 & 18.9 & 71.5 & 47.9 & 61.5 & 35.5 & 20.1 & 9.3  & 23.3 & 10.7 & 0.8  & 6.7  & 21.8 & 12.3 & 43.8 & \textbf{91.6} & 15.0 & 66.7  & 17.3\\
				ResNet-backbone(ours)	& no & yes & \textbf{45.9} & 25.9 & \textbf{73.7} & 58.7 & 53.6 & \textbf{59.0} & \textbf{41.6} & \textbf{30.4} & 15.9 & 12.8 & \textbf{13.8} & \textbf{21.7} & \textbf{47.5} & 31.5 & 71.4 & 87.3 & \textbf{41.1} & \textbf{100.0} & \textbf{40.8}\\
				\bottomrule
			\end{tabular}
		}
		
	\end{center}
	
\end{table*}

\begin{figure*}[!t]
	\centering
	\includegraphics[width=1.0\linewidth]{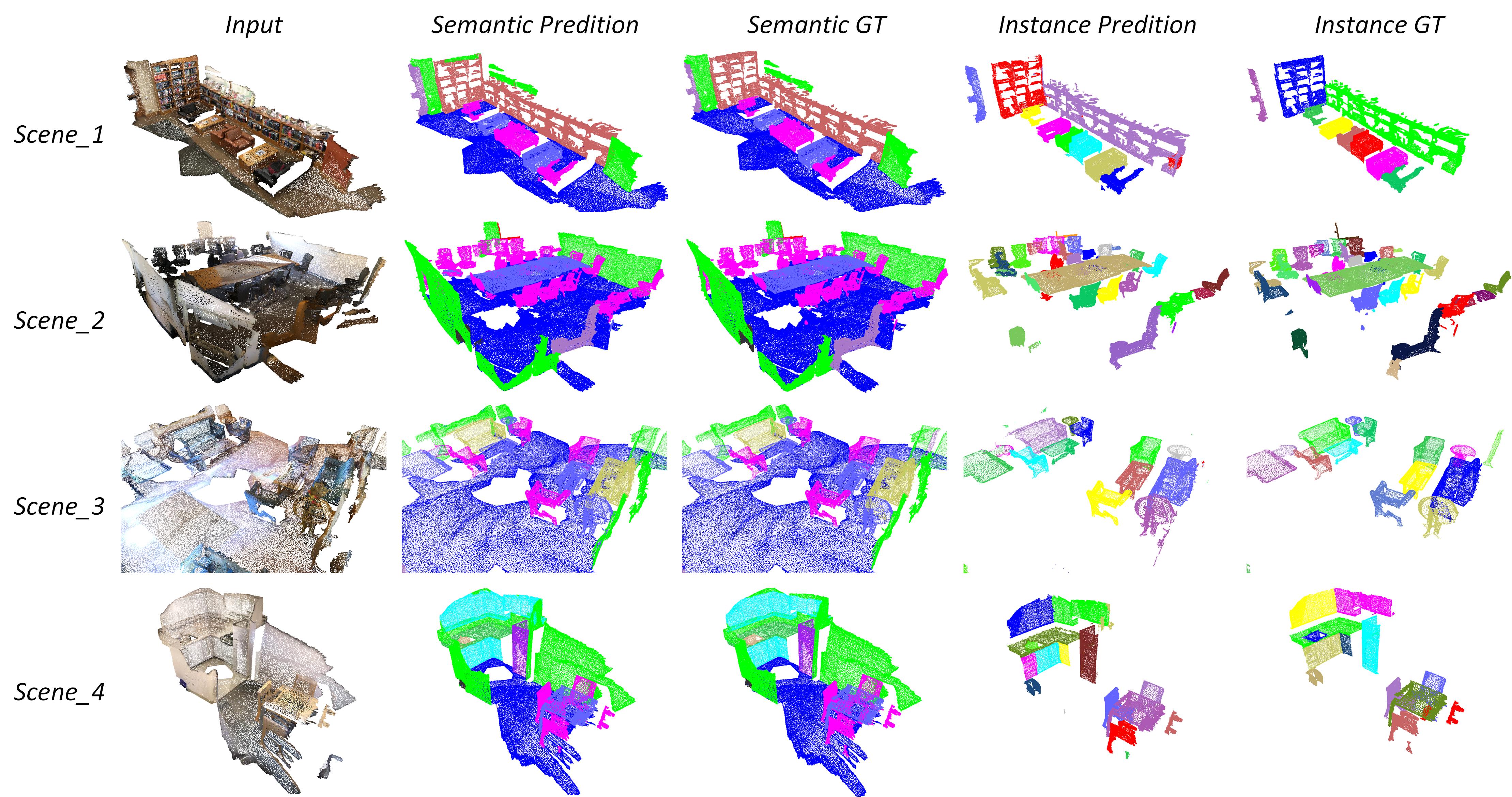}
	\caption{Visualization of ScanNet results. The first column is the input of our model. The second column is the prediction of semantic labels. The third column is the ground truth of semantic segmentation. The forth column is the instance prediction. The fifth column is the ground truth of instance segmentation. For instance segmentation, we only visualize the 18 catagories useful for evaluation while droping other catagories. 
	}
	\label{visualization_scannet}
\end{figure*}

\textbf{Implementation Details. } 
We implement the network with Pytorch1.0 and run it on a single NVIDIA GTX1080Ti. Our network can be easily trained end-to-end. We use the ADAM optimizier with constant learning rate 0.001. $\alpha$ and $\beta$ in the structure-aware loss function is set 0.7 and 1.5 respectively. In our experiment, we use two backbone networks with different model capacities provided by~\cite{graham20183d}. The first backbone network is a UNet-like architecture based on the submanifold sparse convolution with smaller capcity and faster speed. The second is a ResNet-like architecture with larger capcity and slower speed. We train the whole model with the UNet backbone for 50 hours until convergence. The model with the ResNet-backbone needs 150 hours and get much better results. In practice, we pretrain the backbone network first to get a pretrained semantic segmentation model. Then we train the whole model based on the pretrained model. Using pretrained model can save time when conducting multiple experiments.
%What's need to be emphasized is that we do not need to search all points in the scene when using the KNN algorithm. Instead, we just search neighbours within the same semantic catagory and aggregate the information from them. It means that the semantic information provides a mask for KNN. This slight difference decreases the computation cost greatly. 
During the inference, 
%we use the semantic predictions instead of the ground truth as the KNN mask. 
mean-shift algorithm is used to cluster embeddings to get the instance prediction. The bandwidth of mean-shift is set 1.0.

\textbf{Metrics. } 
The average precison (AP) is widely used in the instance segmentation. For ScanNet dataset, the AP with an IoU threshold 0.5 ($AP_{0.5}$) is commonly used. For NYUv2 dataset, the AP with an IoU threshold 0.25 ($AP_{0.25}$) is chosen. For both of the two datasets, images and point clouds (with additional RGB attributes) are provided. Some previous methods use both of the two inputs while others use a single input. In this paper, we only use the point cloud as the input. We argue that we do not use features extracting from images using image-based 2D networks.

\begin{table*}[t]
	
	\caption{3D instance segmentation results on NYUv2 dataset. $AP_{0.25}$ is reported in the table.}
	\label{NYUv2_Result}
	
	\begin{center}
		\resizebox{\textwidth}{!}{
			\begin{tabular}{l|cc|c|ccccccccccccccccccc}
				\toprule
				Method & image & point cloud & mean & bathtub & bed & bookshelf & box & chair & counter & desk & door & dresser & garbage & lamp & monitor & night & pillow & sink & sofa & table & television & toilet\\
				\midrule
				MRCNN & yes & no & 29.3 & 26.3 & 54.1 & 23.4 & 3.1 & 39.3 & 34.0 & 6.2 & 17.8 & 23.7 & \textbf{23.1} & 31.1 & 35.1 & 25.4 & 26.6 & 36.4 & 47.1 & 21.0 & \textbf{23.3} & 58.8 \\
				MRCNN* & yes & no & 31.5 & 24.7 & 66.3 & 20.1 & 1.4 & 44.9 & 43.9 & 6.8 & 16.6 & 29.5 & 22.1 & 29.2 & \textbf{29.3} & 36.9 & 34.6 & 37.1 & 48.4 & 26.6 & 21.9 & 58.5 \\
				SGPN-CNN~\cite{wang2018sgpn} & yes & yes & 33.6 & 45.3 & 62.5 & 43.9 & 0.0 & 45.6 & 40.7 & 30.0 & 20.2 & 42.6 & 8.8 & 28.2 & 15.5 & 43.0 & 30.4 & 51.4 & \textbf{58.9} & 25.6 & 6.6 & 39.0 \\
				R-PointNet-CNN~\cite{yi2018gspn}  & yes & yes & 39.3 & 62.8 & 51.4 & 35.1 & \textbf{11.4} & 54.6 & 45.8 & \textbf{38.0} & 22.9 & \textbf{43.3} & 8.4 & 36.8 & 18.3 & \textbf{58.1} & \textbf{42.0} & 45.4 & 54.8 & \textbf{29.1} & 20.8 & 67.5 \\
				ResNet-backbone(ours)	& no & yes & \textbf{43.0} & \textbf{82.1} & \textbf{67.3} & \textbf{48.1} & 3.5 & \textbf{65.4} & \textbf{56.8} & 14.5 & \textbf{37.6} & 23.1 & 7.3 & \textbf{60.0} & 4.4 & 52.9 & 34.3 & \textbf{68.2} & 55.0 & 28.3 & 20.7 & \textbf{87.2} \\
				\bottomrule
			\end{tabular}
		}
		
	\end{center}
	
\end{table*}

\begin{table}[t]
	
	\caption{Comparison of different network structures.}
	\label{ablation_study}
	
	\begin{center}
		\resizebox{0.5\textwidth}{!}{
			\begin{tabular}{l|cccc}
				\toprule
				Method & AP & AP0.5 & AP0.25 & IoU  \\
				\midrule
				UNet+vanillaLoss & 0.150 & 0.338 & 0.599 & 0.569\\
				UNet+strucLoss & 0.158 & 0.350 & 0.613 & 0.570\\
				UNet+strucLoss+gcn$\times$1 & 0.163 & 0.356 & 0.621 & 0.574\\
				UNet+strucLoss+gcn$\times$2  & \textbf{0.171} & \textbf{0.360} & \textbf{0.630} & 0.572\\
				UNet+strucLoss+gcn$\times$3  & 0.165 & 0.351 & 0.623 & 0.\textbf{576}\\
				\midrule
				ResNet+strucLoss+gcn$\times$2  & \textbf{0.270} & \textbf{0.464} & \textbf{0.672} & \textbf{0.689}\\
				\bottomrule
			\end{tabular}
		}
	\end{center}
	
\end{table}

\subsection{Instance Segmentation on ScanNet}
The ScanNet dataset provides aN online benchmark. So we first evaluate our method on the dataset. 18 categories are used in the instance segmentation task which makes it more challenging task compared to the instance segmentation on a single category. The input of our network is the coordinate and the RGB value of each point.

Our method outperforms all the state-of-arts on the ScanNet benchmark. Among all previous methods, SGPN~\cite{wang2018sgpn} is the most similar method with our method. Compared to SGPN, the space complexity and the computation complexity of our proposed structure-aware function are both $\mathcal{O}(n)$ while these of SGPN are both $\mathcal{O}(n^{2})$. Also, our proposed function considers the structure information while SGPN considers each point equivalently. R-PointNet~\cite{yi2018gspn} and 3D-SIS~\cite{hou20183d} are proposal-based methods. 3D-SIS does not only use the point cloud as the input but also uses images from multiple views. Image information also contributes to its final result. However, our UNet-backbone model and ResNet-backbone model only use the point cloud data as the input.

The UNet-backbone model outperforms almost state-of-arts except 3D-SIS which additionally uses multiple images. The ResNet-backbone model outperforms all methods by a large margin. The result shows the effectiveness of our method. Our method achieves high AP for most categories. Proposal-based methods like R-PointNet and 3D-SIS get higher results for categories such as chairs, sofa, fridges and so on. It is easy to generate bounding boxes for these categories. Mask R-CNN performs better on the picture because the feature of the picture is distinct in 2D images. Our method leverage both semantic information and instance information to generate embeddings for each point so that we can also provide the semantic prediction. Our method is more like the panoptic segmentation instead of just the instance segmentation. Additionally, our method can adapt objects with different sizes and shapes without the limitation of the bounding box.
Qualitative results are showed in Figure~\ref{visualization_scannet}.

\subsection{Instance Segmentation on NYUv2}

Different from ScanNet dataset, NYUv2 dataset provides the single RGBD image instead of the whole scene. Previous methods usually use both images and point clouds as the input to increase the precision on this dataset. We only use the 3D point cloud as the input in this paper. Even though, our method outperforms all state-of-art methods on this dataset. Specially, our method achieves the highest precision for many categories. As NYUv2 dataset provides a single RGBD image with partial point clouds, categories such as boxes, monitors, garbage bins are difficult to recognize only using point clouds. It is easier to segment these categories on the image than on the point cloud. So MRCNN gets better results than our method on some of these categories. Fusing visual features from images can be also helpful. We leave the multi-sensor fusion as a future work.

\subsection{Ablation Study}

We conduct the ablation study on the validation set of the ScanNet (v2) dataset.

\textbf{Different loss functions. }
To validate the effectiveness of our structure-aware loss function, we compare it with a vanilla version loss function. The vanilla version does not use the structure information which means that the importance for each points in the same instance is the same. The first two rows of Table~\ref{ablation_study} show the results. Using structure-aware loss function brings more than 1\% gain on $AP_{0.5}$ and $AP_{0.25}$. Also, we find that the structure-aware loss function does not contribute to the IoU of semantic segmentation. So we think that the increase of the average precision (AP) benefits from more discriminative embeddings supervised by the structure-aware loss function.

\textbf{Different numbers of the GCN layer. }
We compare different numbers of the GCN layer to explore the effectiveness of the GCN. The third to the fifth row in Table~\ref{ablation_study} provides the results with different layer numbers. The GCN layer can contribute to the final result. However, more layers do not mean better result. We find that using two GCN layers achieves the best result. We analyze that more layers may cause overfitting or oversmoothing. This is a common issue for the graph convolution. Also, too many layers may increase the difficulty for training.

\textbf{Different backbone networks. }
Our proposed architecture can adapt to different semantic segmentation backbone networks. In this paper, we compare two models using the UNet backbone and the ResNet backbone respectively. The results are shown in Table~\ref{ablation_study}. The model using the ResNet backbone outperforms the model using the UNet backbone by a large margin. It shows that the backbone network affects our model a lot.

\section{CONCLUSIONS}
We present a novel approach for 3D semantic instance segmentation. The structure-aware loss function considers geometric information to generate discriminative embeddings for instance segmentation. The graph convolutional neural network using attention-based KNN refines intial embeddings by automatic feature selection and aggregation. Experiments show that our approach outperforms all state-of-art methods on ScanNet benchmark and NYUv2 dataset. In the future, multi-sensor fusion can be added to our network to further increase the precision.

\bibliographystyle{named}
\bibliography{bibtex/object_detection,bibtex/instance_segmentation,bibtex/multi_view_representation,bibtex/voxel_representation,bibtex/point_cloud_representation,bibtex/depthwise_convolution,bibtex/graph_cnn,bibtex/others,bibtex/paris,bibtex/dataset}
\end{document}